\documentclass{article}

\usepackage{amsmath}
\usepackage{graphicx}

\usepackage[sglblindworkshop, final]{neurips_2025}

\usepackage[utf8]{inputenc} 
\usepackage[T1]{fontenc}    
\usepackage{hyperref}       
\usepackage{url}            
\usepackage{booktabs}       
\usepackage{amsfonts}       
\usepackage{nicefrac}       
\usepackage{microtype}      
\usepackage{xcolor}         

\title{Unlocking the Invisible Urban Traffic Dynamics under Extreme Weather: A New Physics-Constrained Hamiltonian Learning Algorithm}
\workshoptitle{UrbanAI: Harnessing Artificial Intelligence for Smart Cities}

\author{%
  Xuhui Lin \thanks{Corresponding author for questions about Hamiltonian structure analysis in urban traffic systems.}\\
  The Bartlett School of Sustainable Construction\\
  University College London\\
  London, UK \\
  \texttt{xuhui.lin.16@ucl.ac.uk} \\
  \And
  Qiuchen Lu \\
  The Bartlett School of Sustainable Construction\\
  University College London\\
  London, UK \\
  \texttt{qiuchen.lu@ucl.ac.uk} \\
}

\begin{document}

\maketitle

\begin{abstract}
Urban transportation systems face increasing resilience challenges from extreme weather events, but current assessment methods rely on surface-level recovery indicators that miss hidden structural damage. Existing approaches cannot distinguish between true recovery and "false recovery," where traffic metrics normalize, but the underlying system dynamics permanently degrade. To address this, a new physics-constrained Hamiltonian learning algorithm combining "structural irreversibility detection" and "energy landscape reconstruction" has been developed. Our approach extracts low-dimensional state representations, identifies quasi-Hamiltonian structures through physics-constrained optimization, and quantifies structural changes via energy landscape comparison. Analysis of London's extreme rainfall in 2021 demonstrates that while surface indicators were fully recovered, our algorithm detected 64.8\% structural damage missed by traditional monitoring. Our framework provides tools for proactive structural risk assessment, enabling infrastructure investments based on true system health rather than misleading surface metrics.
\end{abstract}

\section{Introduction}
Urban transportation systems face unprecedented resilience challenges from accelerating urbanization and increasing extreme weather events \cite{chen2022assessing, markolf2019transportation}. Their growing complexity makes them vulnerable to cascading failures and prolonged service disruptions \cite{lin2025field, kouvelas2023cascading}, with restoration processes proving lengthy and expensive \cite{ganin2017resilience, kurth2020lack}. Current resilience assessment methods rely on surface-level recovery indicators such as traffic flow restoration rates and connectivity measures \cite{ poulin2021infrastructure}. These approaches assume that numerical recovery equates to system recovery—if traffic volumes return to pre-event levels, the system is deemed fully restored \cite{besinovic2020resilience}. However, this approach cannot detect whether extreme events have caused permanent structural damage to the system's underlying dynamical architecture, leading to "false recovery" phenomena where surface metrics normalize but fundamental system behavior has been altered. This limitation reflects a broader challenge in complex systems physics, where macroscopic observables may appear stable while underlying dynamical structures have been permanently changed \cite{arvidsson2023urban, battiston2023higher}. Current resilience quantification approaches lack calibration across different stress scenarios \cite{martins2024assessing, esmalian2022quantitative}, and empirical observations during disasters remain limited \cite{zhang2024resilience}.

Recent advances in physics-informed machine learning offer new possibilities for understanding complex system dynamics. Hamiltonian Neural Networks can learn conservation laws from data while maintaining physical consistency \cite{greydanus2019hamiltonian, mattheakis2022hamiltonian}, and port-Hamiltonian approaches show promise for modeling structural dynamics with energy-preserving properties \cite{desai2021port, roth2025stable}. However, existing algorithms face critical gaps when applied to urban transportation systems: 1) structural invisibility—current methods cannot detect hidden changes in underlying energy landscapes after extreme events\cite{cuturi2022understanding, wang2021physics, latrach2023critical}, 2) false recovery detection failure—no algorithm can distinguish between true recovery and cases where surface metrics normalize but fundamental architecture remains altered\cite{sun2020resilience, besinovic2020resilience}, and 3) multi-scale integration deficiency—existing approaches cannot simultaneously capture both quasi-Hamiltonian conservation properties and dissipative effects across temporal scales\cite{mattheakis2022hamiltonian, desai2021port}. To address these limitations, we introduce "structural irreversibility" for urban transportation systems—the phenomenon where surface indicators recover while the system's underlying dynamical architecture remains permanently altered. Unlike traditional resilience metrics focusing on functional recovery \cite{liu2025resilience, zhou2021improving}, we develop a novel Physics-Constrained Hamiltonian Structure Learning (PCHSL) algorithm that learns the system's intrinsic "energy landscape" to diagnose structural health changes invisible to conventional monitoring. We validate this framework through analysis of London's transportation network during the October 5, 2021 extreme rainfall, demonstrating that while surface indicators fully recovered, the system's Hamiltonian structure underwent significant and persistent changes.
\section{Methods}

\subsection{Problem Formulation}
Traditional resilience assessment methods face a fundamental limitation: they monitor surface phenomena (flow rates, speeds, connectivity) but cannot detect changes in the underlying organizational principles that govern system behavior \cite{sun2020resilience, zhang2024resilience}. This is analogous to observing that a ball returns to a similar position without knowing whether the landscape it rolls on has been permanently altered. We develop a novel PCHSL algorithm to solve this structural invisibility problem. Our algorithm models urban traffic systems as dynamical systems with discoverable energy-based structures. Transportation networks exhibit conservation-like properties (flow conservation at intersections) and natural equilibrium-seeking behavior (stable traffic patterns), suggesting they may follow quasi-Hamiltonian dynamics \cite{greydanus2019hamiltonian, desai2021port}. The core innovation of PCHSL algorithm is learning the "energy landscape"--the underlying structural rules that determine how the system evolves and responds to disturbances \cite{mattheakis2022hamiltonian}. Given traffic time series data $\mathbf{X} \in \mathbb{R}^{N \times T}$ from $N$ road segments, we hypothesize the system follows:
\begin{equation}
\dot{\mathbf{z}} = \mathbf{J} \nabla H(\mathbf{z}) + \mathbf{D}(\mathbf{z})
\end{equation}
where $\mathbf{z} \in \mathbb{R}^d$ represents the system's low-dimensional state, $H(\mathbf{z})$ is the Hamiltonian function encoding the energy landscape, $\mathbf{J} = \begin{bmatrix} 0 & 1 \\ -1 & 0 \end{bmatrix}$ is the symplectic matrix preserving energy structure, and $\mathbf{D}(\mathbf{z})$ captures dissipative effects from friction and control interventions. The Hamiltonian $H(\mathbf{z})$ acts as a "topographic map" where valleys represent stable operating states and hills represent unstable configurations. The system naturally evolves toward energy minima (stable traffic patterns), following gradients of this landscape. Crucially, the PCHSL algorithm detects when extreme events alter the landscape structure–changing the locations, depths, or shapes of these valleys–indicating that the system's fundamental behavior is permanently modified, even if surface metrics appear to recover. Our algorithm reveals such structural irreversibility by comparing energy landscapes before and after extreme events, uncovering hidden changes that traditional monitoring cannot capture.

\subsection{The PCHSL Algorithm: A Three-Module Architecture}
The PCHSL algorithm addresses the challenge that real transportation systems operate in high-dimensional space, making direct Hamiltonian analysis intractable. The algorithm consists of three core computational modules: \textbf{Module 1} applies PCA\cite{mackiewicz1993principal} and t-SNE\cite{maaten2008visualizing} to extract essential dynamics: $\mathbf{z}_t = \mathbf{U}_d^T \mathbf{x}_t$, where $\mathbf{U}_d$ contains the first $d=2$ principal components. \textbf{Module 2} implements our core algorithmic innovation: physics-constrained optimization to learn the Hamiltonian structure. Our algorithm parameterizes the Hamiltonian as: $H(z_1, z_2) = \sum_{i,j} h_{ij} z_1^i z_2^j$ and compute derivatives using finite differences: $\dot{z}_i(t) = \frac{z_i(t+\Delta t) - z_i(t-\Delta t)}{2\Delta t}$. The learning objective enforces Hamiltonian structure:
\begin{equation}
    \mathcal{L}(\boldsymbol{\theta}) = \frac{1}{T} \sum_{t=1}^T |\dot{\mathbf{z}}_t - \mathbf{J} \nabla H(\mathbf{z}_t; \boldsymbol{\theta})|^2 + \lambda \mathcal{R}(\boldsymbol{\theta})
\end{equation}
\textbf{Module 3} provides uncertainty quantification through Bayesian inference with Gaussian priors and Hamiltonian Monte Carlo sampling for robust parameter estimation of the learned energy landscape.

\subsection{Structural Irreversibility Detection}
The central hypothesis of our PCHSL algorithm is that extreme events may permanently alter the energy landscape while leaving surface traffic patterns apparently unchanged. To test this hypothesis, we learn separate Hamiltonians $H_{\text{before}}$ and $H_{\text{after}}$ for pre- and post-event periods, then systematically compare their structural properties. We quantify structural differences by measuring how much the energy landscapes have changed across the relevant state space:
\begin{equation}
    d(H_{\text{before}}, H_{\text{after}}) = \int_{\Omega} |H_{\text{before}}(\mathbf{z}) - H_{\text{after}}(\mathbf{z})|^2 d\mathbf{z}
\end{equation}
This metric captures fundamental changes in the "topographic map": whether valleys have shifted, deepened, or become shallower, regardless of whether traffic flows have returned to similar aggregate levels. To enable meaningful comparison across different systems and events, we normalize this distance to obtain the structural irreversibility index
\begin{equation}
    \text{SIR} = \frac{d(H_{\text{before}}, H_{\text{after}})}{\|H_{\text{before}}\|_2}
\end{equation}
This threshold reflects significant landscape deformation that would measurably affect system stability and future response patterns.

\section{Results}
\begin{figure}[htbp]
    \centering
    \includegraphics[width=1\textwidth]{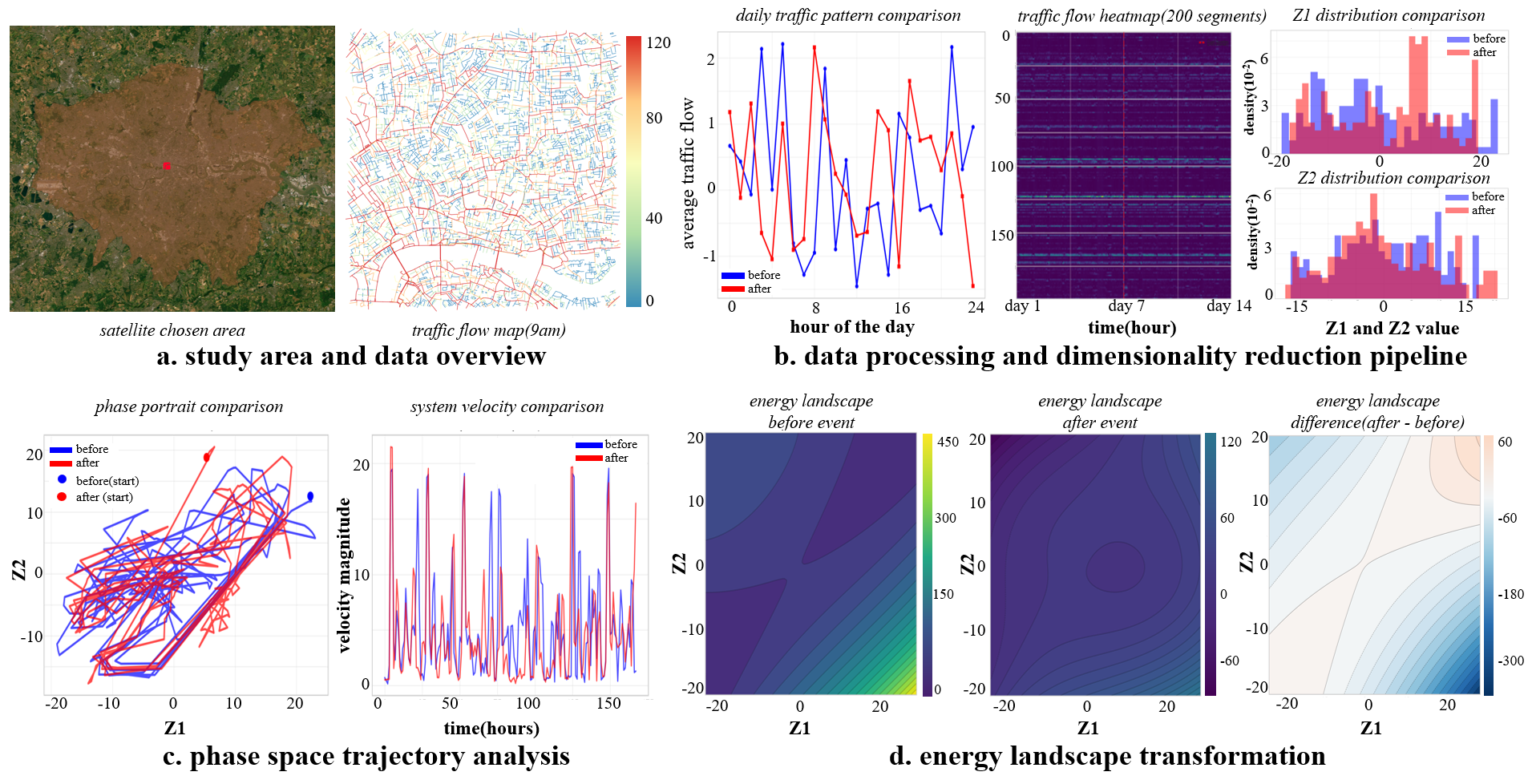}
    \caption{Structural irreversibility detection in London's transportation system under extreme rainfall}
    \label{fig:diagram1}
\end{figure}

We analyze London's transportation network during the extreme rainfall event of October 5, 2021, using traffic flow data from 11,821 road segments collected at hourly intervals. The dataset spans 14 days (7 days before and 7 days after the event), totaling 3,971,856 data points with rainfall intensity of 42 mm/h (Fig.\ref{fig:diagram1}a). Our data processing pipeline demonstrates effectiveness in capturing the system's underlying dynamics (Fig.\ref{fig:diagram1}b). The daily traffic pattern comparison reveals apparent recovery in surface-level metrics, with both pre- and post-event periods showing similar diurnal patterns. However, the traffic flow heatmap indicates subtle but persistent changes in temporal-spatial distribution patterns. Our two-stage dimensionality reduction (PCA to 50 dimensions retaining 77.5\% variance, then t-SNE to 2D) successfully captured essential system dynamics. Our physics-constrained optimization identifies quasi-Hamiltonian structures, achieving 92.96\% and 98.81\% convergence improvement for pre- and post-event periods respectively. The phase space trajectory analysis (Fig.\ref{fig:diagram1}c) reveals striking differences in system behavior. The pre-event trajectories show more dispersed patterns with broader exploration of the state space, while post-event trajectories exhibit more constrained dynamics with clear convergence toward specific regions. The system velocity comparison demonstrates reduced dynamical variability in the post-event period, suggesting the emergence of more regular but fundamentally different organizational patterns. The learned energy functions reveal profound structural transformation. The pre-event Hamiltonian takes the form:
$$H_{\text{before}}(z_1, z_2) = 0.045 - 0.095z_2 + 0.149z_2^2 - 0.046z_1 - 0.236z_1z_2 + 0.058z_1^2$$
while the post-event function becomes:
$$H_{\text{after}}(z_1, z_2) = 0.070 + 0.082z_2 + 0.044z_2^2 - 0.192z_1 + 0.002z_1z_2 - 0.008z_1^2$$
The energy landscape transformation (Fig.\ref{fig:diagram1}d) provides visual evidence of structural irreversibility. The before-event energy landscape shows a complex topography with multiple local minima and smooth gradients. In contrast, the after-event landscape exhibits a fundamentally different structure with altered valley locations and modified energy barriers. Most critically, the difference map reveals systematic shifts in energy distribution, with some regions experiencing energy increases up to +60 units while others show decreases down to -300 units. Despite apparent surface-level recovery in daily traffic patterns, our structural analysis reveals significant hidden damage. The energy landscape distance $d(H_{\text{before}}, H_{\text{after}})$ yields a Structural Irreversibility Index of $\text{SIR} = 0.695$, exceeding our detection threshold by nearly an order of magnitude. The coefficient sign changes in critical terms indicate fundamental alterations in dynamical coupling mechanisms. Our framework estimates a 64.8\% false recovery component, demonstrating that conventional monitoring approaches miss approximately two-thirds of actual structural damage. The phase space analysis further confirms this structural transformation. While pre-event trajectories exhibit continuous flux behavior without stable attractors, the post-event system develops defined convergence patterns toward specific state space regions. This represents a qualitative phase transition from a system characterized by broad exploration to one with constrained dynamics, providing quantitative evidence that extreme weather events can induce permanent organizational transitions in urban traffic systems that remain completely invisible to traditional surface-level monitoring approaches.

\section{Discussion}
The proposed algorithm provides the first quantitative evidence that extreme weather events can cause structural irreversibility in urban transportation systems, fundamentally challenging the assumption that surface-level recovery indicates true system restoration. The algorithm's detection of 64.8\% false recovery component represents a critical blind spot in current resilience assessment practices, revealing that traditional metrics systematically miss nearly two-thirds of actual structural damage. Our algorithm's identification of coefficient sign changes and energy landscape transformation provides converging computational evidence for genuine dynamical regime change. Most significantly, the algorithm successfully detects the emergence of stable equilibria, indicating a qualitative phase transition in system self-organization, with implications for future perturbation responses. Our physics-constrained algorithmic framework demonstrates the value of embedding conservation principles into machine learning, providing mechanistic insight into structural changes invisible to conventional monitoring approaches. The concept of structural irreversibility, enabled by our algorithm, may extend to other critical infrastructure systems where surface metrics similarly mask permanent organizational changes. Algorithm limitations include single-event analysis and quasi-Hamiltonian assumptions. Future algorithmic developments should validate across diverse contexts and develop real-time monitoring capabilities.

\bibliography{references}

@article{chen2022assessing,
  title={Assessing and enhancing urban road network resilience under rainstorm waterlogging disasters},
  author={Chen, Xin and Miller-Hooks, Elise and Karimi, Ali},
  journal={Transportation Research Part D: Transport and Environment},
  volume={109},
  pages={103355},
  year={2022},
  publisher={Elsevier}
}

@article{cuturi2022understanding,
  title={Understanding and Mitigating Gradient Flow Pathologies in Physics-Informed Neural Networks},
  author={Cuturi, Marco and Wang, Sifan and Berman, Scott and Perdikaris, Paris},
  journal={SIAM Journal on Scientific Computing},
  volume={44},
  number={5},
  pages={A3055--A3081},
  year={2022},
  publisher={SIAM}
}

@article{markolf2019transportation,
  title={Transportation resilience to climate change and extreme weather events--Beyond risk and robustness},
  author={Markolf, Samuel A and Hoehne, Christopher and Fraser, Anne and Chester, Mikhail V and Underwood, B Shane},
  journal={Transportation Research Part A: Policy and Practice},
  volume={121},
  pages={376--389},
  year={2019},
  publisher={Elsevier}
}

@article{wang2021physics,
  title={Physics-informed machine learning: A survey on problems, methods and applications},
  author={Wang, Shengze and Teng, Yuting and Perdikaris, Paris},
  journal={Computer Methods in Applied Mechanics and Engineering},
  volume={385},
  pages={114084},
  year={2021},
  publisher={Elsevier}
}

@article{latrach2023critical,
  title={A Critical Review of Physics-Informed Machine Learning Applications in Subsurface Energy Systems},
  author={Latrach, Abdeldjalil and Marzouk, Youssef and Eldridge, Matthew and Voskov, Denis and Hajibeygi, Hadi},
  journal={arXiv preprint arXiv:2308.04457},
  year={2023}
}

@article{lin2025field,
  title={Field-theory Inspired Physics-Informed Graph Neural Network for Reliable Traffic Flow Prediction under Urban Flooding},
  author={Lin, Xuhui and Lu, Qiuchen and Zhao, Pengjun and Chen, Long and Tang, Junqing and Guan, Dabo and Broyd, Tim},
  journal={Reliability Engineering \& System Safety},
  pages={111487},
  year={2025},
  publisher={Elsevier}
}

@article{kouvelas2023cascading,
  title={Cascading failures and resilience evolution in urban road traffic networks with bounded rational route choice},
  author={Kouvelas, Anastasios and Leclercq, Ludovic and Geroliminis, Nikolas},
  journal={Physica A: Statistical Mechanics and its Applications},
  volume={612},
  pages={128481},
  year={2023},
  publisher={Elsevier}
}

@article{ganin2017resilience,
  title={Resilience and efficiency in transportation networks},
  author={Ganin, Alexander A and Kitsak, Maksim and Marchese, Dayton and Keisler, Jeffrey M and Seager, Thomas and Linkov, Igor},
  journal={Science Advances},
  volume={3},
  number={12},
  pages={e1701079},
  year={2017},
  publisher={American Association for the Advancement of Science}
}

@article{kurth2020lack,
  title={Lack of resilience in transportation networks: Economic implications},
  author={Kurth, Moritz and Kozlowski, Witold and Ganin, Alexander and Mersky, Avi and Leung, Brian and Dykes, Jason and Kitsak, Maksim and Linkov, Igor},
  journal={Transportation Research Part A: Policy and Practice},
  volume={138},
  pages={63--81},
  year={2020},
  publisher={Elsevier}
}

@article{poulin2021infrastructure,
  title={Infrastructure resilience curves: Performance measures and summary metrics},
  author={Poulin, Cynthia and Kane, Michael B},
  journal={Reliability Engineering \& System Safety},
  volume={216},
  pages={107926},
  year={2021},
  publisher={Elsevier}
}

@article{sun2020resilience,
  title={Resilience metrics and measurement methods for transportation infrastructure: the state of the art},
  author={Sun, Weiwei and Bocchini, Paolo and Davison, Brian D},
  journal={Sustainable and Resilient Infrastructure},
  volume={5},
  number={3},
  pages={168--199},
  year={2020},
  publisher={Taylor \& Francis}
}

@article{besinovic2020resilience,
  title={Resilience in railway transport systems: a literature review and research agenda},
  author={Besinovic, Nikola},
  journal={Transport Reviews},
  volume={41},
  number={4},
  pages={457--478},
  year={2020},
  publisher={Taylor \& Francis}
}

@article{arvidsson2023urban,
  title={Urban scaling laws arise from within-city inequalities},
  author={Arvidsson, Martin and Lovsjo, Niclas and Keuschnigg, Marc},
  journal={Nature Human Behaviour},
  volume={7},
  pages={365--374},
  year={2023},
  publisher={Nature Publishing Group}
}

@article{battiston2023higher,
  title={Higher-order interactions shape collective dynamics differently in hypergraphs and simplicial complexes},
  author={Battiston, Federico and Cencetti, Giulia and Iacopini, Iacopo and Latora, Vito and Lucas, Maxime and Patania, Alice and Young, Jean-Gabriel and Petri, Giovanni},
  journal={Nature Communications},
  volume={14},
  pages={1605},
  year={2023},
  publisher={Nature Publishing Group}
}

@article{martins2024assessing,
  title={Assessing transport network resilience: empirical insights from real-world data studies},
  author={Martins, Maria Cecilia M and Rodrigues da Silva, Antonio Nelson and Pinto, Nuno},
  journal={Transport Reviews},
  volume={44},
  number={5},
  pages={834--857},
  year={2024},
  publisher={Taylor \& Francis}
}

@article{esmalian2022quantitative,
  title={Quantitative measures for integrating resilience into transportation planning practice: Study in Texas},
  author={Esmalian, Armin and Yuan, Fang and Dong, Shangjia and Karamlou, Ali and Mostafavi, Ali},
  journal={Transportation Research Part A: Policy and Practice},
  volume={164},
  pages={140--157},
  year={2022},
  publisher={Elsevier}
}

@article{zhang2024resilience,
  title={Resilience analysis of maritime transportation networks: a systematic review},
  author={Zhang, Dan and Tao, Jianfeng and Wan, Chengpeng and Huang, Liang and Yang, Molin},
  journal={Transportation Safety and Environment},
  volume={6},
  number={4},
  pages={tdae009},
  year={2024},
  publisher={Oxford University Press}
}

@article{greydanus2019hamiltonian,
  title={Hamiltonian neural networks},
  author={Greydanus, Sam and Dzamba, Misko and Yosinski, Jason},
  journal={Advances in Neural Information Processing Systems},
  volume={32},
  pages={15353--15363},
  year={2019}
}

@article{mattheakis2022hamiltonian,
  title={Hamiltonian neural networks for solving equations of motion},
  author={Mattheakis, Marios and Sondak, David and Protopapas, Pavlos and Ribeiro, Marcelo},
  journal={Physical Review E},
  volume={105},
  pages={065305},
  year={2022},
  publisher={American Physical Society}
}

@article{desai2021port,
  title={Port-Hamiltonian neural networks for learning explicit time-dependent dynamical systems},
  author={Desai, Shaan A and Mattheakis, Marios and Sondak, David and Protopapas, Pavlos and Roberts, Stephen J},
  journal={Physical Review E},
  volume={104},
  pages={034312},
  year={2021},
  publisher={American Physical Society}
}

@article{roth2025stable,
  title={Stable Port-Hamiltonian Neural Networks},
  author={Roth, Florian J and Schaller, Manuel and Worthmann, Karl and Ober-Blobaum, Sina and Eckstein, Berkay},
  journal={arXiv preprint arXiv:2502.02480},
  year={2025}
}

@article{liu2025resilience,
  title={Resilience modeling of transportation infrastructure and network based on the semi-Markov process considering resource dependency},
  author={Liu, Zihao and Xie, Jingcheng},
  journal={Reliability Engineering \& System Safety},
  volume={253},
  pages={110512},
  year={2025},
  publisher={Elsevier}
}

@article{zhou2021improving,
  title={Improving the resilience of urban transportation to natural disasters: the case of Changchun, China},
  author={Zhou, Yang and Wang, Jiaoe and Yang, Haibo},
  journal={Scientific Reports},
  volume={14},
  pages={26253},
  year={2021},
  publisher={Nature Publishing Group}
}

@article{mackiewicz1993principal,
  title={Principal components analysis (PCA)},
  author={Ma{\'c}kiewicz, Andrzej and Ratajczak, Waldemar},
  journal={Computers \& Geosciences},
  volume={19},
  number={3},
  pages={303--342},
  year={1993},
  publisher={Elsevier}
}

@article{maaten2008visualizing,
  title={Visualizing data using t-SNE},
  author={Maaten, Laurens van der and Hinton, Geoffrey},
  journal={Journal of machine learning research},
  volume={9},
  number={Nov},
  pages={2579--2605},
  year={2008}
}
\end{document}